\documentclass[journal]{IEEEtran}
\usepackage[utf8]{inputenc}
\usepackage[english]{babel}
\usepackage[dvipsnames,table,xcdraw]{xcolor}
\usepackage{multirow}
\usepackage{float}
\usepackage{graphics} 
\usepackage{graphicx}
\usepackage{multicol}
\usepackage{subcaption}
\usepackage{color, colortbl}
\usepackage{amsmath} 
\usepackage{array}  
\usepackage{algorithm}
\usepackage[noend]{algpseudocode}

\makeatletter
\def\BState{\State\hskip-\ALG@thistlm}
\makeatother

\begin{document}
%
\title{Long-term map maintenance pipeline for autonomous vehicles}
%
%
%

\author{Julie Stephany Berrio,~\IEEEmembership{Member,~IEEE,} 
        Stewart Worrall,~\IEEEmembership{Member,~IEEE,}\\
        Mao Shan,~\IEEEmembership{Member,~IEEE,}
        and Eduardo Nebot,~\IEEEmembership{Member,~IEEE}
        
\thanks{J. Berrio, S. Worrall, M. Shan and E. Nebot are with the 
Australian Centre for Field Robotics (ACFR) at the University of Sydney, NSW, Australia. E-mails: {\tt \{j.berrio, s.worrall, m.shan e.nebot\}@acfr.USyd.edu.au}.}
\thanks{Manuscript received on April XX, 2020; revised on Month XX, 2020.}}

%
%

\markboth{Journal of \LaTeX\ Class Files,~Vol.~14, No.~8, August~2015}%
{Shell \MakeLowercase{\textit{et al.}}: Bare Demo of IEEEtran.cls for IEEE Journals}
%



\maketitle

\begin{abstract}

For autonomous vehicles to operate persistently in a typical urban environment, it is essential to have high accuracy position information. This requires a mapping and localisation system that can adapt to changes over time. A localisation approach based on a single-survey map will not be suitable for long-term operation as it does not incorporate variations in the environment.
In this paper, we present new algorithms to maintain a featured-based map. A map maintenance pipeline is proposed that can continuously update a map with the most relevant features taking advantage of the changes in the surroundings. 
Our pipeline detects and removes transient features based on their geometrical relationships with the vehicle's pose. Newly identified features became part of a new feature map and are assessed by the pipeline as candidates for the localisation map. By purging out-of-date features and adding newly detected features, we continually update the prior map to more accurately represent the most recent environment.  
We have validated our approach using the USyd Campus Dataset \cite{Usyd_dataset} \cite{USYD_Segmentation_2019}, which includes more than 18 months of data. The results presented demonstrate that our maintenance pipeline produces a resilient map which can provide sustained localisation performance over time.

\end{abstract}


%
\IEEEpeerreviewmaketitle

\section{Introduction}

\IEEEPARstart{A}{n} autonomous vehicle (AV) software stack has four main components: perception, localisation, planning/control, and mapping. 
Perception provides information about the environment in proximity to the AV. Localisation provides the position of the vehicle within a local or global context. Planning and control is responsible for the planning of trajectories, decision making and control algorithms. Finally, mapping is a requirement for each of the previous components \cite{hd_maps}. The localisation system requires an accurate map to determine the vehicle's position with centimetre-level accuracy. The map can be also used for locating different dynamic objects detected by the perception system to plan the future vehicle manoeuvres and share this information with other vehicles. 
For these reasons, maps are a fundamental component required for the development and deployment of AVs. Building and maintaining the maps to centimetre-level accuracy is essential for the reliability and safety of the system.

\begin{figure}[t]
\centering
    \includegraphics[width=\columnwidth]{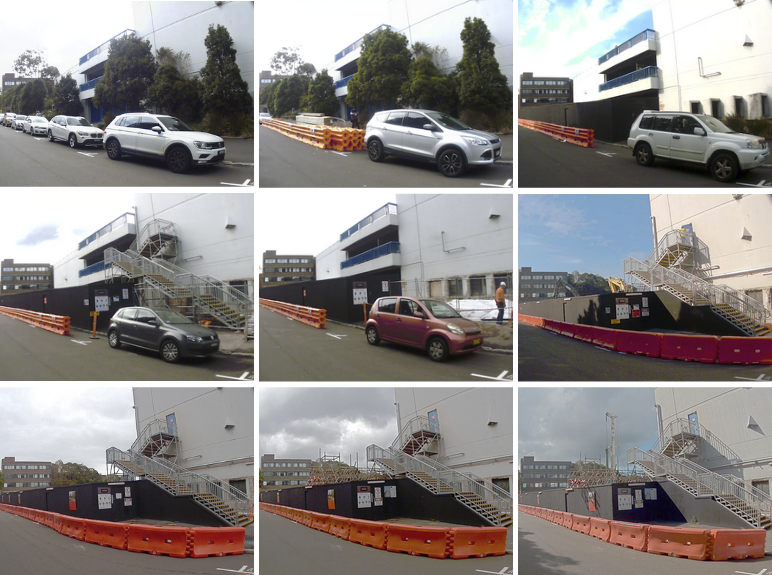}
    \caption{\small Evolution of a construction site (closure, demolition and building work) recorded in the Usyd Campus Dataset.}
    \label{fig:construction_site}
\end{figure}

Before AVs are deployed to public streets, a mapping process must be performed by estimating the location of unique features (within a reference frame) which describe the driving environment. 
The map must be detailed enough to allow the localisation algorithm to match the observations and identify the vehicle's precise pose within it. Map building remains an arduous task due to a lack of standardization for the representation of the map, the amount/quality of data required, the cost of operating data-collection platforms, and the processing time.  

\begin{figure}[t]
\centering
    \includegraphics[width=\columnwidth]{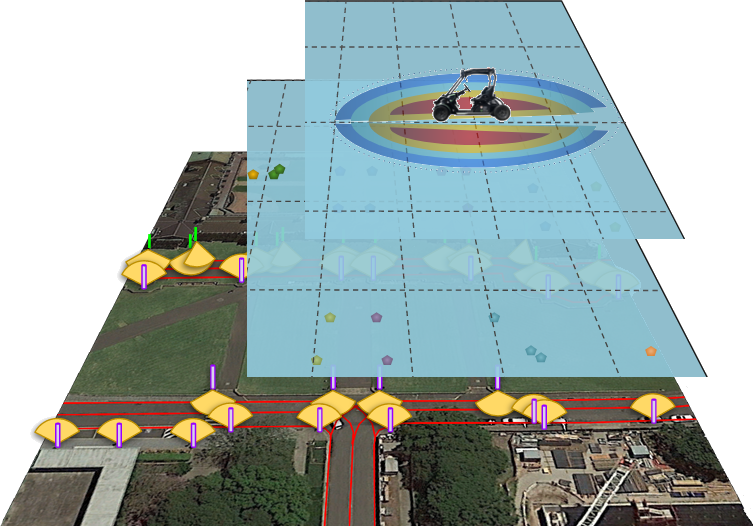}
    \caption{\small Layers for long-term map maintenance. The prior map layer at the bottom contains the output of the mapping process, the features' location within a global frame and the location from which they were detected. The new feature map layer in the middle consists of newly detected features and their characteristics. The sensor model layer on the top describes how the vehicle's sensor system identifies the different features. }
    \label{fig:map_levels}
\end{figure}

For most AVs, computation power and storage capacity is limited, making the selection of the map representation important for real-time system performance. Different mapping companies such as Tomtom \cite{tomtom} with "RoadDNA" and CivilMaps  \cite{civilmaps} with "Fingerprint maps" have achieved a significant reduction in the amount of data required to perform map matching and maintain an acceptable localisation accuracy. These maps are built using a distinctive and detectable pattern of features for the urban driving environment. It includes poles, traffic signals, road markings and/or voxel-based fingerprints. Feature-based maps for autonomous driving purposes can be as lightweight as 100 KB per km. In this way, the maps could be stored on-board, shared, and downloaded from the cloud efficiently using wireless communication.

One of the main challenges in using maps for autonomous driving applications is that a map created from data collected on a given day might not be entirely valid the next week. This is due to potential changes in the environment such as construction sites as shown in Fig. \ref{fig:construction_site} or other temporary structures.

Maps require continuous updates because unexpected changes to the environment could cause problems with localisation, or other system components, potentially resulting in an accident. It is essential to have algorithms capable of incorporating the environmental changes into the map. There are two standard approaches to implement the map maintenance processes. One is based on cumulative adjustment; actions that lead to preserve the original features while adding new features in response to the changes. This approach has two main disadvantages. Since all new features are added to the map, it grows without limit. At the same time, this makes the data association more difficult between the observations and the map. The second approach is based on transformative adjustments; operations that modify the initial map structure to incorporate environmental variations. Besides adding new features into the map, this approach also deals with the removal of transient map components.
The main difference between these two methodologies is the deletion of transitory features. While both approaches store all of the newly detected features, the second approach also removes features that no longer exist making it more appropriate in terms of managing the map size and simplifying the data association task.

There are different approaches that are currently capable of solving one of the two problems under certain conditions; removal of transient features or addition of new features into the map. Nevertheless, there are only a small number of state-of-the-art approaches which can handle both tasks. To the best of our knowledge, none of these can be easily adapted to different localisation algorithms or can operate independently, meaning that the mapping pipeline requires information from the data association and the vehicle's pose which are generic in any map-based localisation system.  

\begin{figure}[b]
\centering
    \includegraphics[width=0.65\columnwidth]{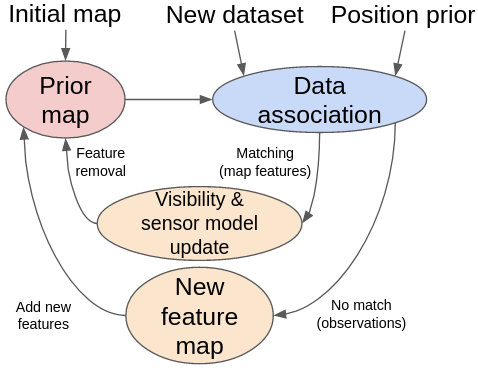}
    \caption{\small High-level flowchart of the process.}
    \label{fig:journal_process_high_level}
\end{figure}

The proposed pipeline aims to solve both problems; removal of transient features and addition of newly detected features. We present a layered-map approach to storing information about the detected and matched features, as shown in Fig. \ref{fig:map_levels}. Our approach incorporates a pipeline that works in conjunction with the localisation algorithm, as shown in Fig. \ref{fig:journal_process_high_level}. In the prior map layer, we have a feature-based localisation map in a global reference frame in which we update the features' properties to detect and remove transient features. 

A generic localisation system is usually composed of two modules: a data association module in charge of identifying the correspondence between the observations and the map, and an estimation algorithm which calculates the a posteriori state of the vehicle based on different sensors and the output of the previous module.
%
%
The data association module can be adjusted to retrieve three pieces of information; the successfully matched and unmatched map features, and the observations which were detected but not associated to corresponding features in the current map. This last piece of information is stored and evaluated in the new feature map layer of our pipeline to then add potentially valuable features for localisation to the prior map.

The sensor model layer of our pipeline is a 2D grid centred in the vehicle's coordinates frame. This model depicts the probability of the feature extractor detecting different landmarks in the area around the vehicle.
The proposed pipeline uses different algorithms and data structures to store, retrieve and update each of the layers.  
The pipeline is independent of the localisation algorithm used. It will only require information about the current vehicle pose, matched and non-matched map features, and non-matched observations. 
This paper proposes a comprehensive and novel probabilistic pipeline for feature-based map maintenance which addresses the removal of transient features and inclusion of new detected features. The major contributions in this paper are:
\begin{itemize}
  \item A grid-based methodology to detect and remove transient features in a localisation map.
  \item A procedure to include newly detected features into the localisation map.
  \item A demonstration of the two previous contributions in the form of a pipeline, capable of modifying a prior map based on changes to the environment. This is demonstrated using a long term dataset.
\end{itemize}

We demonstrate capabilities of the proposed algorithms using the USyd Campus Dataset \cite{Usyd_dataset} and a localization algorithm which includes a prior feature-based map, a localization filter and a data association method.
The paper is organised as follows: In the next section, we present the state-of-the-art in map maintenance. In Section III, we present the proposed algorithms for the long-term map maintenance pipeline. 
In Section IV, we explain the dataset, the initial map and the localisation algorithm used to test and evaluate our pipeline. 
We demonstrate the performance of the algorithms by experiments with the results presented in Section V. In Section VI, we present further discussions about the results. Finally, the conclusion and future work are presented in Section VII.

\section{RELATED WORK}

A map is required to perform accurate global localisation in an urban environment where GNSS is not always available.
A feature based map can be built using landmarks which best represent the operational environment, also considering the memory and computation resources of the platform. 
The map also needs to be updated as the environment changes; adding new information into the map when discovered, and rely on data association/matching algorithms to reject information that is no longer valid (cumulative adjustments). On the other hand, a more complex approach is to remove invalid landmarks, and incorporate new landmarks into the map as the changes are detected (transformative adjustments). 

Churchill and Newman in \cite{Churchill2012} presented plastic maps for lifelong navigation. This method adds distinct visual experiences to the map when there is a change in the scene (represented as a failure in localisation). The different experiences are linked by places, which are created when the agent can be localised in more than one experience. 
In \cite{Einhorn2013} authors describe a graph-based SLAM \cite{Kretzschmar2010} approach which was designed to allow merging of new map fragments to the localisation map. MacTavish et al. in \cite{Barfoot_2018} present a methodology to recall relevant experiences in visual localisation, the multi-experience localisation system stores every visual experience in a layer and selects the one more suitable for the current view.

In turn, the transformative adjustments can be classified into two groups: those which only perform removal of features, and those which also simultaneously add new information into the map.
Amongst the first approaches, we can find algorithms that evaluate the temporal properties of each feature. Biber and Duckett in \cite{Biber2005, Biber2009} represented the environment using multiple and discrete timescales, where depending on the semantic category of each feature, a vanishing rate time is established. Later, Rosen et al. in \cite{Rosen2016} presented a probabilistic model based on the feature persistence that illustrates the survival time of each map component. In \cite{Nobre2018}, the authors extended \cite{Rosen2016} by adding a module able to capture the potential correlation between features, allowing modelling their persistence jointly.
In \cite{stephany2019_2}, we exploited the geometric connections between vehicle's pose over several drives and the observation of features to define the visibility of landmarks. Changes in the area of visibility, specifically reduction (caused by non-detection), leads to the removal of the feature. 

Rating and selection of features is another approach used for long-term localisation. This methodology ranks the landmarks based on their value for localisation. The score function of each feature depends on different predictors.
Hochdorfer and Schlegel in \cite{Hochdorfer2009} clustered landmarks that covered approximately the same observation area, and calculated the information content of each landmark base on its uncertainty. In order to remove landmark, each cluster is assessed and the ones with the lowest localisation benefit are deleted. 

An approach based on scoring the features by the number of detection was presented in \cite{Muhlfellner2016}. The authors chose this particular predictor due to its correlation with the average number of matches in different datasets. Nevertheless, this particular predictor is susceptible to factors like vehicle speed and areas visited. In \cite{Dymczyk2015},  authors used a relation between the number of observed trajectories and its expectancy. This work was expanded in \cite{dymczyk2016} where the authors combined the predictor as: distance travelled while observing a landmark, mean re-projection error and the classification of the descriptors appearance as ranking function. 
A similar approach was presented in \cite{stephany2019_1}. The ranking scheme consisted of a regression model that was trained with predictors describing the density of landmarks within the map and the way they were detected by the sensor system. 

Some approaches have demonstrated persistent long-term localisation by selecting the appropriate features to create the map. In \cite{carneiro2009}, the authors assess the stability of visual features by exploiting their distinctiveness and the robustness of image descriptors. In \cite{Verdie2015}, the uniqueness of the descriptors is evaluated to determine whether features should be included into the map.
A learning-based approach to select features that are robust to varied lightning conditions has been implemented in \cite{Lategahn2013} to achieve long-term localisation.  

Methodologies which allow the addition and deletion of features to the map are more suitable for automotive applications as this can minimise the size of the map without affecting the localisation accuracy.
Konolige and Bowman in \cite{Konolige2009} presented a view-based map derived from FrameSLAM \cite{Pirker2011}. This approach allows the inclusion of new features while deleting views. This last step is performed by calculating the percentage of the number of matches of each view and deleting the least-recently used (LRU) views. 
In \cite{Egger2018}, Egger et al. chose nodes 3D surfels to represent the environment. A C-SLAM algorithm is used to integrate new nodes to replace those which had changed. 

During the last few years, new approaches related to the periodicity of features have been published. These works assume that some of the detected and mapped landmarks are detected at a predetermined frequency in the environment.
This is the case of \cite{Krajnik2017}, Krajnik et al. introduced FreMEn, where the probability of the map elements' states are represented by a combination of functions with different amplitudes and frequencies which represents the changes in the environment. The authors extended their previous work in \cite{Krajnik2019} by using a spatial-temporal representation to calculate the probabilistic distribution of the features in time. 

A number of the long-term techniques used for visual localisation requires the storing of different versions for the same place, and identify which one is the most suitable for the current circumstances in order to localise within it. These techniques are particularly suited to localisation with sensors such as cameras, in which measurements change over the course of a day due to time dependent lighting conditions. In this paper, we assume the map features are illumination invariant, e.g., lidar/radar features. 

The solution of the map maintenance problem has become an essential requirement for the deployment of AVs. In \cite{Pannen_2020}, the authors present an approach to detect changes in HD maps \cite{Pannen_2019} on highways and consequently replace identified zones (based on the nature of localisation algorithm) with lane marking patches to keep the map updated. In contrast, our feature-based map is suitable for urban environments where the environment is more structured and where lane markings cannot always be detected on the road. 

So far, we have presented a discussion about different approaches to map maintenance which are capable of removing transient features, and others that add new features into the map. Nevertheless, in the literature, there are no existing approaches which run completely independently of the localiser used while addressing all these issues together. The majority of these works are specific to the implementation of the localisation algorithm or filter to be able to perform the map updates.
In the following section, we present the proposed methodology to update feature-based maps, followed by modules that are used to develop and evaluate the pipeline. 

\section{MAP MAINTENANCE METHODOLOGY}

\begin{figure}[b]
\centering
    \includegraphics[width=\columnwidth]{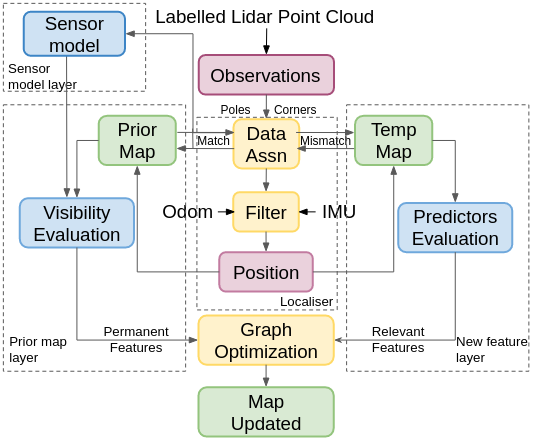}
    \caption{ \small Flow chart of the process. }
    \label{fig:map_flowc}
\end{figure}

In this section, we present a novel pipeline to update a prior feature-based map by removing transient elements and adding newly detected features.
The proposed pipeline consists of three map layers. The prior map layer represents the localisation layer consisting of the prior map. This layer is updated by making use of map matching information generated by the data association algorithm. Through the maintenance of this layer, we aim to detect which features are no longer existent and to remove them from the localisation map.
The new feature map layer represents a structure where we save and update all observations which have not been associated with an existing map element. 

In a parallel process, the current observations are also compared against this map to identify the components which are recurrently observed. 
The sensor model layer corresponds to a grid map which stores lidar detections of features around the vehicle. 
The prior and new feature map layer are assessed in order to keep or remove features from the respective layers. The permanent and relevant features form the prior and new feature map layer, respectively, and are ultimately merged together in a new and updated map through an optimization algorithm. 
Fig. \ref{fig:map_flowc} shows a flow chart which contains the main processes and their connections to perform the maintenance of the map.
Each element of the long-term pipeline will be explained in the following subsections.


\subsection{Sensor model layer}

\begin{figure}[t]
\centering
    \begin{subfigure}[b]{0.68\columnwidth}
    \centering
     \includegraphics[width=\columnwidth]{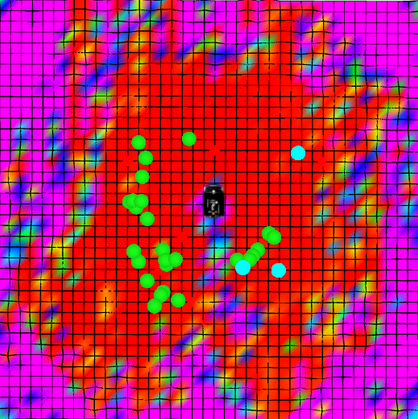}
  \end{subfigure}
    \begin{subfigure}[b]{0.052\columnwidth}
    \centering
    \textbf{1\\}
    \includegraphics[width=0.92\textwidth]{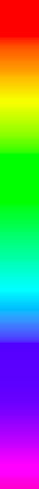}
    \textbf{0}
  \end{subfigure}
\caption{\small Sensor model; the grid map is coloured by the probability of each cell to detect a feature within it: \colorbox{Red}{red} colour represents a probability of 1.0 while \colorbox{Magenta}{magenta} 0.0 probability of  detecting a corner \textcolor{cyan}{$\bullet$} or a pole \textcolor{green}{$\bullet$}. }
\label{fig:sensory_model}    
\end{figure}

The sensor model layer represents how the sensor system perceives the features from its surroundings. It consists of a 2D grid which is centred on the vehicle's coordinate frame as shown in Fig. \ref{fig:sensory_model}. In our experiments, the size of the grid is set to 60 meters; selected due to the capabilities of the lidar to detect features of interest at longer ranges. For the case of the Velodyne VLP-16 lidar, poles and corners are generally detected within a 30 meters radius of the vehicle.
Every time a detection or misdetection of a feature occurs, the cell corresponding to the location of the feature in the vehicle frame is updated. The update is implemented using a binary Bayes filter. The value in each cell $l_{ct}$ denotes a log-odds representation of the probability of observing a feature within that region.

 \begin{equation} \label{eq0}
l_{ct}(v|z_{1:t})=l_{ct-1}(v|z_{1:t-1}) + l_c(v|z_{t}) 
\end{equation}

In our experiments, we set log-odds values of $l_c(v|z_{t}) =0.7$ for detection and $l_c(v|z_{t})=-0.4$ for misdetection.

\subsection{Prior map layer}

\begin{figure}[h]
\centering
 \begin{subfigure}[]{\columnwidth}
    \centering
     \includegraphics[width=0.95\columnwidth]{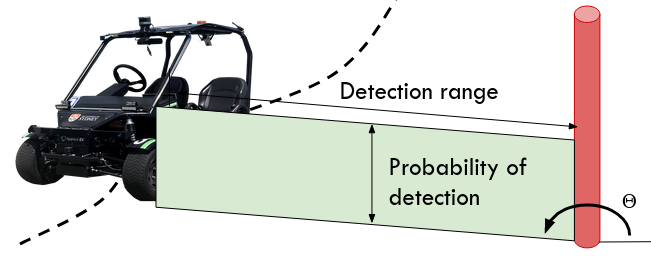}
     \caption{\small Single pair detection range - probability of detection. }
     \label{fig:pair_vis}
  \end{subfigure}
  
 \begin{subfigure}[]{0.9\columnwidth}
    \centering
     \includegraphics[width=0.9\columnwidth]{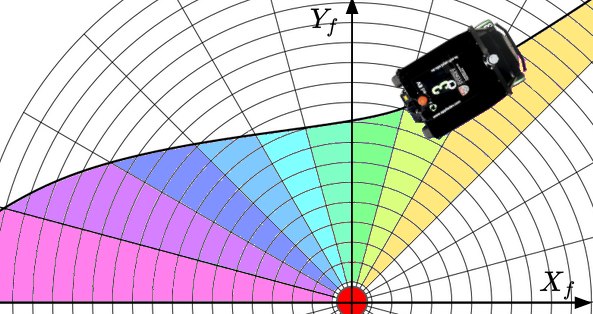}
     \caption{\small Depiction of the detection range visibility vector. }
     \label{fig:traj}
  \end{subfigure}
    \begin{subfigure}[]{0.04\columnwidth}
    \centering
    \includegraphics[width=\textwidth]{Figures/KSEy7.png}
    \caption{ }
    \label{fig:color}
  \end{subfigure}
\caption{\small Depiction of the visibility vectors for one single feature (in red). Fig. \ref{fig:pair_vis} illustrates a the detection range and the probability of detection for one discrete angle $\theta$. Fig. \ref{fig:traj} shows the trajectory of the vehicle, the area under the curve corresponds to the region from where the feature can be detected by the vehicle. 
}
\label{fig:map_vis}
\end{figure}

This layer is composed of the localisation map complemented by a measure of the visibility of each feature. The visibility of a feature is defined by two vectors. Each element of the vectors corresponds to a discretised angle between 0 and 360 $^\circ$. The first vector contains the maximum range from where the feature was detected at a particular angle. 
Fig. \ref{fig:map_vis} shows a representation of the range vector for a single feature. 
The second vector consists of the probability of detection at that particular angle. The number of elements in the vector depends on the angular resolution.

\begin{figure*}[h]
\centering
\begin{subfigure}[]{\textwidth}
    \centering
    \includegraphics[width=0.98\textwidth]{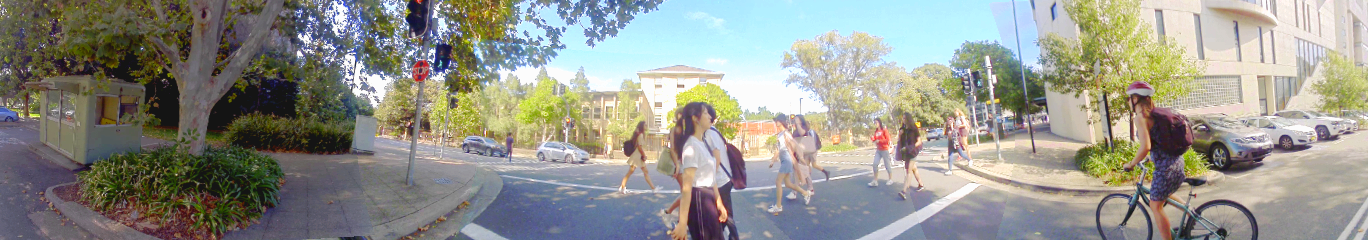}
    \caption{\small Stitched images for a 350$^\circ$ field of view of the surroundings.}
    \label{fig:enviroment}
  \end{subfigure}

    \begin{subfigure}[]{0.493\columnwidth}
    \centering
    \includegraphics[width=0.98\textwidth]{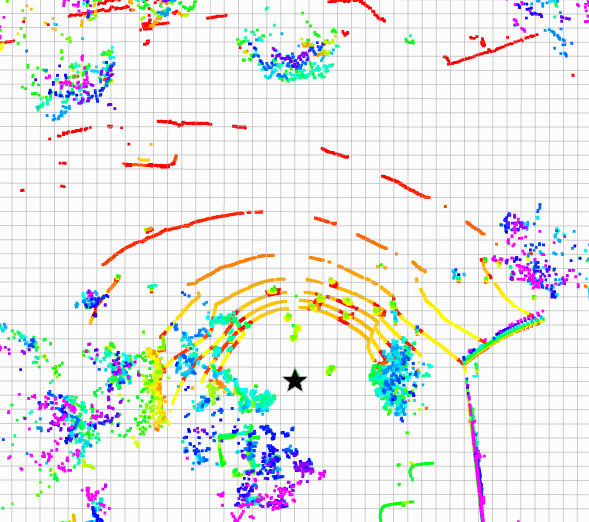}
    \caption{\small Point cloud}
    \label{fig:lidar_observ}
  \end{subfigure}
    \begin{subfigure}[]{0.493\columnwidth}
    \centering
    \includegraphics[width=0.98\textwidth]{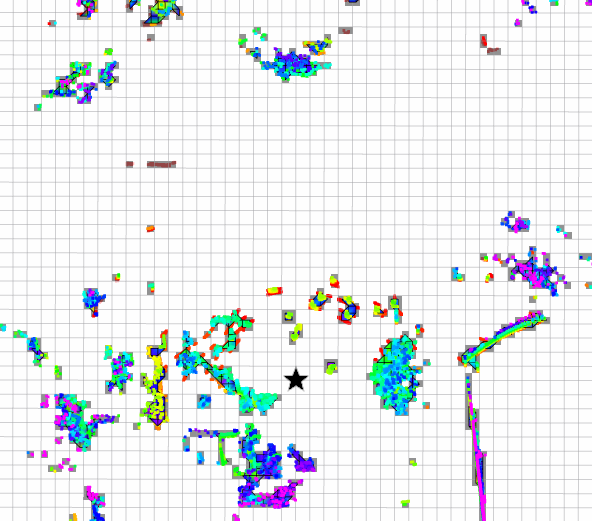}
    \caption{\small 2D grid obstacles and PC }
    \label{fig:lidar_obs}
  \end{subfigure}
  \begin{subfigure}[]{0.493\columnwidth}
  \centering
    \includegraphics[width=0.98\textwidth]{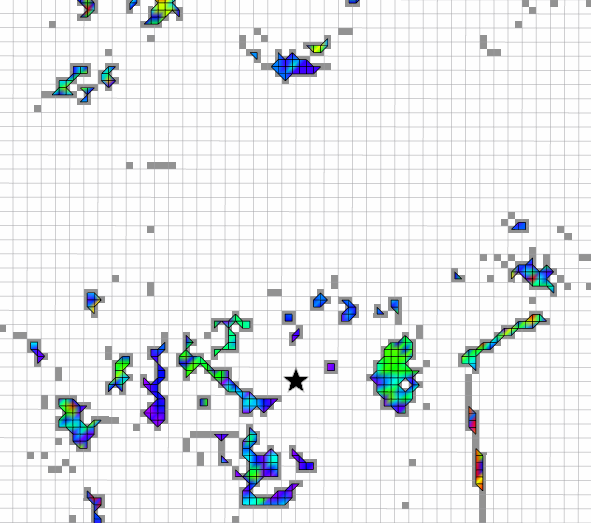}
    \caption{\small 2D grid obstacles}
    \label{fig:grid}
  \end{subfigure}
  \centering
  \begin{subfigure}[]{0.493\columnwidth}
    \includegraphics[width=0.98\textwidth]{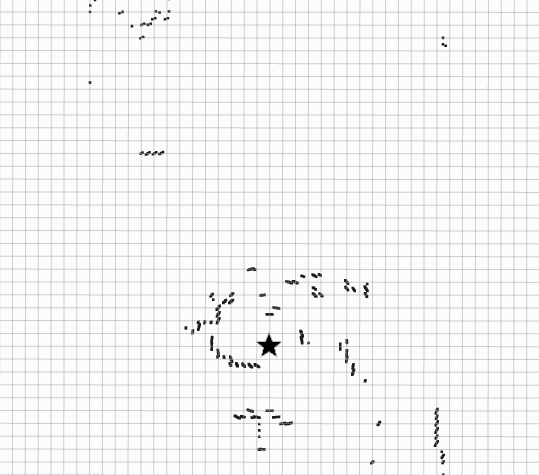}
    \caption{\small Ray casting }
    \label{fig:raycast}
  \end{subfigure}
\caption{\small Occlusion detection process. The reference image of the surroundings is shown in Fig. \ref{fig:enviroment}. Fig. \ref{fig:lidar_observ}, \ref{fig:lidar_obs}, \ref{fig:grid} and \ref{fig:raycast} depict the point cloud occlusion handling process where the vehicle is represented by a  black star facing in the upwards orientation.}
\label{fig:occlusion}
\end{figure*}

The range vector is updated each time the feature is matched with the observations and the current detection range for the same angle is larger than the previous one. In that case, the element of the vector will take the current range value. A particular element of the range vector will also be updated if the following three conditions are met: the map feature has not been matched, the corresponding range in the vector is larger than current and the feature is not occluded by any obstacle. 



The second vector represents the visibility and corresponds to the detection probability of each feature $l_{pt}$ at specific angle. Similar to the update of the range vector, one cell of the vector is updated with either a match, or when a feature is both not detected and not occluded. We use the log-odds representation of the Bayes filter to perform this task:

\begin{equation} \label{eq6}
l_{pt}=l_{pt-1} \pm \left |l_{ct}(v|z_{1:t})\right |
\end{equation}

The $l_{pt}$ is given by its previous value plus (for detections) or minus (for misdetections) $l_{ct}$, which is the log-odds value of the corresponding cell in the sensor model layer.

The data association algorithm outputs both the successfully matched, and non-matched features. This method does not consider if the feature is occluded. Urban environments are often characterized by having many dynamic objects; vehicles on the road or pedestrians on the sidewalks. These non-static objects could partially or completely occlude the localisation features, and this occlusion could be permanent or temporary depending on the nature of the object. 

To check if a particular non-matched feature was occluded or not, we consider the objects in the environment by evaluating the line of sight from the vehicle to the feature.
We detect the obstacles in the environment by maintaining a 2.5 D grid map \cite{Fankhauser2016GridMapLibrary} centred in the vehicle's footprint frame. 
Initially, the surrounding environment is divided into a grid of $r$ x $r$ meters  and $n$ x $n$ cells. The lidar point cloud is transformed into the footprint frame and then projected to the grid map. The highest and lowest 3D points in the z-axis are obtained for every cell of the grid map. We then compute the difference between the maximum and minimum value in the z-axis. Those cells in which the height difference is greater than a threshold $th_{gm}$, are considered to be occupied by an obstacle. 

Once we have obtained the obstacles from the environment in a grid map format, we perform a 2D ray casting. The origin of all rays is fixed relative to the obstacle grid, and it corresponds to centre point of the vehicle's frame. The ray casting algorithm generates an output similar to a 2D laser scan, where a range value is associated with each discrete angle.
Since we know the position of the features with respect to the vehicle's frame, we can establish if there is an obstacle between the lidar and the non-detected features. The features in the vehicle's frame are converted from Cartesian to polar coordinates and then compared with the ray casting algorithm output to check for occlusion.
Fig. \ref{fig:occlusion} illustrates the methodology used to identify obstacles between the vehicle and the features detected. 

Verifying for occlusion will allow us to perform a more reliable update of the visibility of the map features. The Algorithm \ref{visibility_update} shows a summary of the process for updating the visibility vector representing a particular vehicle pose. 

\begin{algorithm} [t]
\caption{Visibility Update}\label{visibility_update}
\begin{algorithmic}[1]
\Procedure{Visibility Vectors Update}{}
\State $\textit{map\_features} \gets [X_f, Y_f]$
\State $\textit{vehicle\_pose} \gets (x_v, y_v)$
\State $\textit{number\_features} \gets \text{size}(\textit{map\_features})$
\State $i \gets 0$
\While{$i<\textit{number\_features} $}  
\State $angle \gets \text{atan2}\left ( (y_v - y_{f_i})\mathbin{/}(x_v - x_{f_i})\right )$
\State $disc\_angle \gets \text{round}(angle*180\mathbin{/}\pi)+180$ 
\State $range \gets \sqrt{(y_v - y_{f_i})^2 + (x_v - x_{f_i})^2}$
\If {Matched}
\If {$Ranges(disc\_angle) < range$}
\State $Ranges(disc\_angle) \gets range$.
\State $l_{pt}(disc\_angle)=l_{p_{t-1}}(disc\_angle) + l_{ct}$
\EndIf
\ElsIf{non-Occluded}
\If {$Ranges(disc\_angle) > range$}
\State $Ranges(disc\_angle) \gets range$ -1 .
\State $l_{pt}(disc\_angle)=l_{p_{t-1}}(disc\_angle) - l_{ct}$
\EndIf
\EndIf
\State $i \gets i+1$.
\EndWhile
\EndProcedure

\end{algorithmic}
\end{algorithm}

By assessing the evolution of the visibility vector, we can select the transient features which will be removed from the map. We calculate the visibility volume as: 

\begin{equation} \label{eq2}
V_f = \sum_{\alpha =0}^{\alpha =360} 0.5*range(\alpha)^2 *P_{pt}(\alpha),
\end{equation}
where: 
 \begin{equation} \label{eq8}
P_{pt}=1-\frac{1}{1+e^{l_{pt}}}.
\end{equation}

We calculate the difference between the visibility area before and after each drive of the dataset. A reduction in visibility for a particular feature implies that the feature is no longer observable. It means that a feature which was not occluded could no longer be detected by the lidar given that it was observed before from the same angle and a larger or same range.
If the reduction in the visibility value is greater than a threshold $th_{vis}$, the feature is removed from the map. We set the threshold according to \cite{stephany2019_2} where $12\%$ showed the best performance. 

\subsection{New feature map layer}

The new feature map layer stores the observations of new features which have not been matched against the prior map. New features can appear due to a structural change of the environment, or because they were occluded in the previous data that was used to create the prior map. In our experiments, for the first month of operation we set the minimum height parameters of the feature detector the same as the ones used to build the map. The map building process is especially sensitive to the features' parameters, the values of 1.8 m and 1.6 m for each type of features ensure the successful loop closure, noise reduction and optimization.


After building the initial map, for usual operation and localisation purposes, we loosen the constraints of the feature detection algorithm. A consequence of these changes is that new features can be detected which were previously out of the range given the previous height threshold limit. With this change, the existing features can also be observed from a larger range, benefiting the data association algorithm and leading to improved localisation accuracy.

Typically, these newly observed features would be discarded due to not being already included in the prior map, meaning they would not be matched.
In this paper, we propose to build an alternate, new feature map to store this recent information. When the localisation algorithm is initialised, the detected observations which are not matched to the prior map are transformed into the global frame, and included in the new feature map layer.


The new mismatched observations are compared against the features in the new feature map layer, for this purpose we use an extra iterative closest point (ICP) based data association and a Euclidean clustering algorithm. 
This information is needed to calculate evaluation parameters for each feature, besides the final building a graph to optimize the new features' position.

\begin{figure}[h]
\centering
    \begin{subfigure}[b]{0.75\columnwidth}
    \centering
    \includegraphics[width=\textwidth]{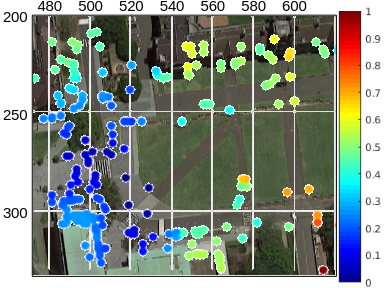}
    \caption{Concentration ratio}
    \label{fig:cr}
  \end{subfigure}
  
    \begin{subfigure}[b]{0.75\columnwidth}
    \centering
    \includegraphics[width=\textwidth]{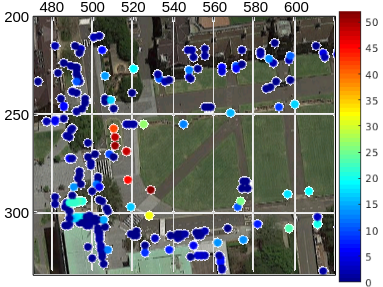}
    \caption{\small Distance travelled }
    \label{fig:variables_layer1}
  \end{subfigure}
\caption{\small Variables involved in the selection of features from the new feature map layer. In Fig. \ref{fig:cr} the density of features is higher where the concentration ratio is lower (in blue) and vice-versa on the red features. In Fig. \ref{fig:variables_layer1} the distance travelled while detecting the feature is greater than 40m for the features coloured in red, while shorter than 5 meters for the dark blue ones. }
\label{fig:predictors}
\end{figure}

At the end of each week, we performed Euclidean clustering of all observations from the new feature map layer. 
In case the ICP data association algorithm fails, observations which belong to a cluster in which standard deviation is lower than 15 cm are linked together. 


We demonstrate that it is possible to quantify the localisation relevance of a feature by using different predictors \cite{stephany2019_1}. This requires the output of the data association process and the vehicle location with respect to the features frame of reference. We used two predictors to assess the quality of the new features to determine whether they should be included in the localisation map.

For each timestep, we obtain the data about the geometric relationship between the vehicle pose and the observed features. With this information, we calculate metrics such as the distance travelled while the feature can be observed, and the ratio describing the concentration of features within the localisation map.

The measure of distance travelled while detecting a given feature is calculated by monitoring the set of vehicle poses from where a particular feature was observed. The importance of the feature increases if the feature can be observed over a longer distance; this would result in more opportunities to update the vehicle pose within the localisation framework.
The concentration ratio on the other hand, is a measurement of the distribution of the features.

\begin{equation} \label{eq4}
\textrm{Concentration ratio} =  \frac{max(df_{n})}{\sum{df_{n}}}
\end{equation}

The concentration ratio is calculated per feature and is defined as the ratio between the distance to the furthest landmark $max(df_{n})$, and the sum of the distances $df_{n}$ to all other features within a specified range. The lower this value is, the higher the density of features is. A concentration ratio approaching one indicates that the landmark density is sparse.
With this variable, we look to prioritize the inclusion of new features where the distribution is low. 

\begin{figure}[t]
\centering
    \includegraphics[width=\columnwidth]{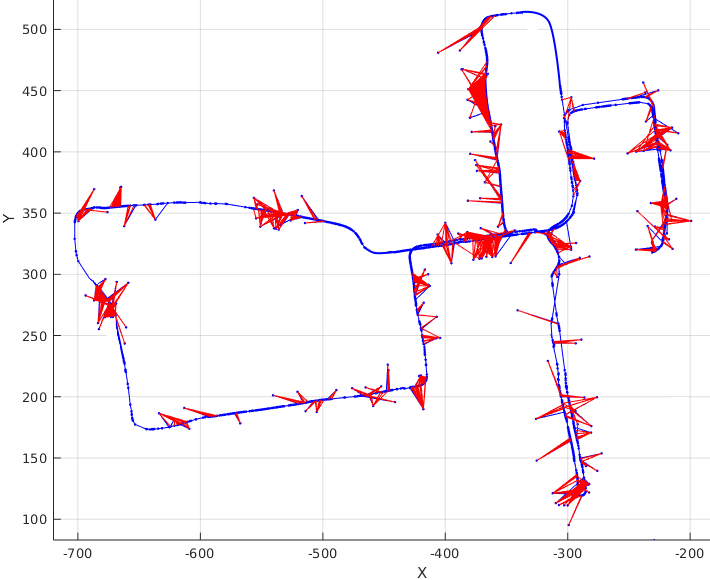}
    \caption{\small Optimized pose graph of new detected features.}
    \label{fig:map_nodes}
\end{figure}

We select the features from the new feature map layer that was visible to the vehicle for more than one-metre distance. We found that for our data, one-meter distance allows us to include corners which are important for localisation but not often detected. 
From this we construct a graph with nodes are composed of feature locations and edges from the vehicle's poses. 
The graph is then optimized to adjust the coordinates of the features as shown in Fig. \ref{fig:map_nodes}.
For each of the resulting features, the concentration ratio is calculated within the map. Those features which have a score of less than 0.4 (belonging to a high-feature dense area) are discarded since we don't want overcrowded regions in the map (making the map more prone to data association mistakes), instead, a more distributed feature map. Then, the remaining features are merged into the prior map layer.

\section{MAP EVALUATION METHODOLOGY}

In the previous section we presented the methodology for a pipeline capable of updating a featured-based map. 
In this section, we present three components used to analyze and evaluate the long-term mapping solution. 
The first element is a dataset comprising of an extensive set of repeated vehicle trajectories with an array of sensor readings collected over a long period of time. 
The time span of the dataset covers structural changes in areas that were repeatedly visited. This component is essential to evaluate the changes in the environment over time.

The second component corresponds to an initial localization map created with the first set of data, during a single drive around the area where no special traffic conditions were enforced. This map provides us with a starting point to test our pipeline.

The third element is a localisation algorithm capable of computing the global pose of the vehicle within the prior map based on the available sensor data.

\subsection{Dataset}

The evaluation dataset was obtained with an electric vehicle (EV) retrofitted with multiple sensors, and covers a timespan of more than 18 months.  The weekly trip covered approximately the same route around the University of Sydney campus. The dataset includes diverse weather/environmental conditions, variation in illumination, the creation and removal of structures and buildings, and diversity in the volume of pedestrian and vehicle traffic.   

\begin{figure}[h]
\centering
    \includegraphics[width=0.92\columnwidth]{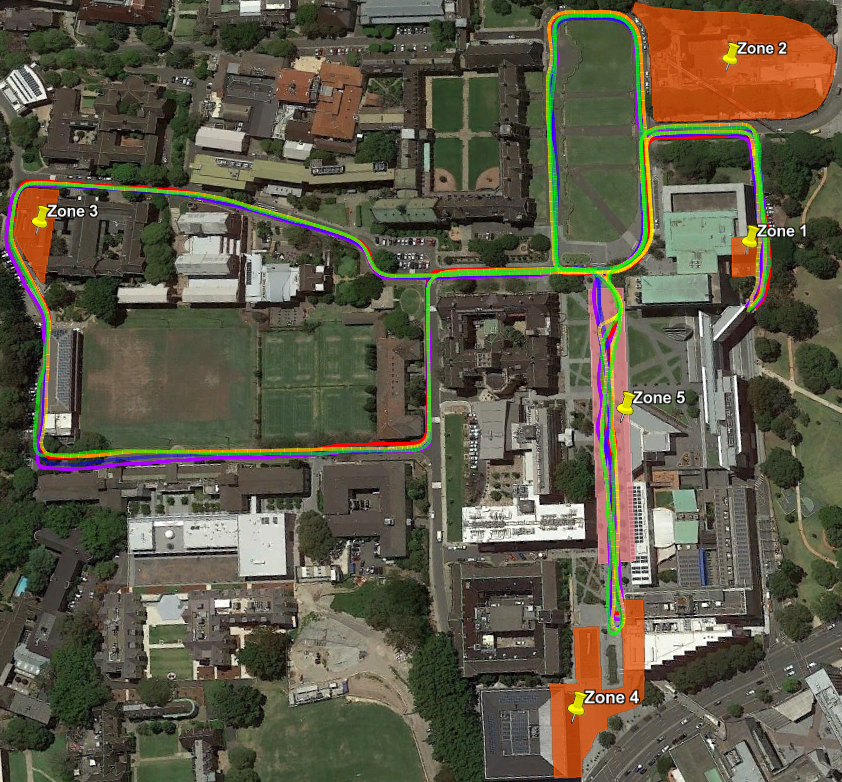}
    \caption{\small The partial trajectory of the University of Sydney (USyd) campus dataset used in this work. The highlighted zones 1-4 in \colorbox{Orange}{orange} correspond to urban development and construction sites (renovations to facades, new building constructions, shut-downs, etc.) which changed considerably over the course of the data collection period. Zone 5, on the other hand, is predominantly a pedestrian space, where frequent university events often result in changes to the surroundings.}
    \label{fig:dataset_track}
\end{figure}

While driving, we logged the readings from various sensor modalities sensing, which allowed us to develop and test different algorithms (perception, localisation, navigation, etc). Fig. \ref{fig:dataset_track} shows the subset of the trajectories from the dataset used in this paper (this portion has roughly the same quality of GNSS data). 

The EV used for the dataset is equipped with:

\begin{itemize}
	\item One Velodyne Puck LITE - 16 beam lidar sensor placed on the vehicle's front roof (tilted $7^\circ$ downwards), it provides 3D information of the environment in form of point cloud at a frequency of 10 Hz. 
	\item Six NVIDIA 2Mega SF3322 automotive GMSL cameras mounted around the vehicle for a complete $360^\circ$ coverage of the vehicle's surroundings. Images were recorded at 30 frames per second with 2.3 M pixel resolution. 
	\item One VN-100 Inertial Measurement Unit (IMU) which computes and outputs real-time 3D orientation. 
	\item Four wheel incremental encoders built into the motors, which allow the speed estimation of each individual wheel.
	\item One potentiometer joined to the steering wheel to calculate the front wheels' steering angles.
	\item One U-Blox NEO-M8P GNSS module which specify the geo-spatial positioning in longitude, latitude, and altitude coordinates.
\end{itemize}

\begin{figure}[h!]
\centering
    \includegraphics[width=0.85\columnwidth]{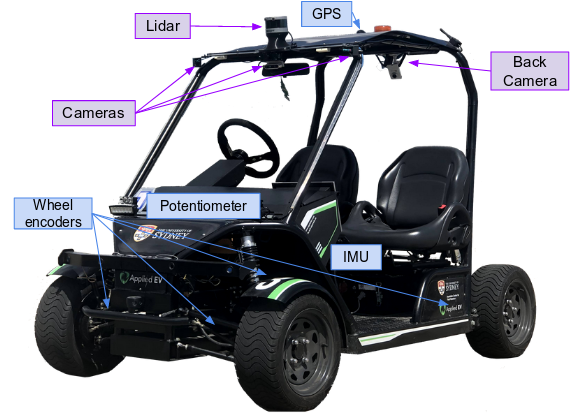}
    \caption{\small Sensor arrangement on the data collection platform. }
    \label{fig:car_sensor}
\end{figure}

The location of the different sensor modalities on the collection platform is depicted in Fig. \ref{fig:car_sensor}. 

\begin{figure*}[!ht]
\centering
    \includegraphics[width=0.95\textwidth]{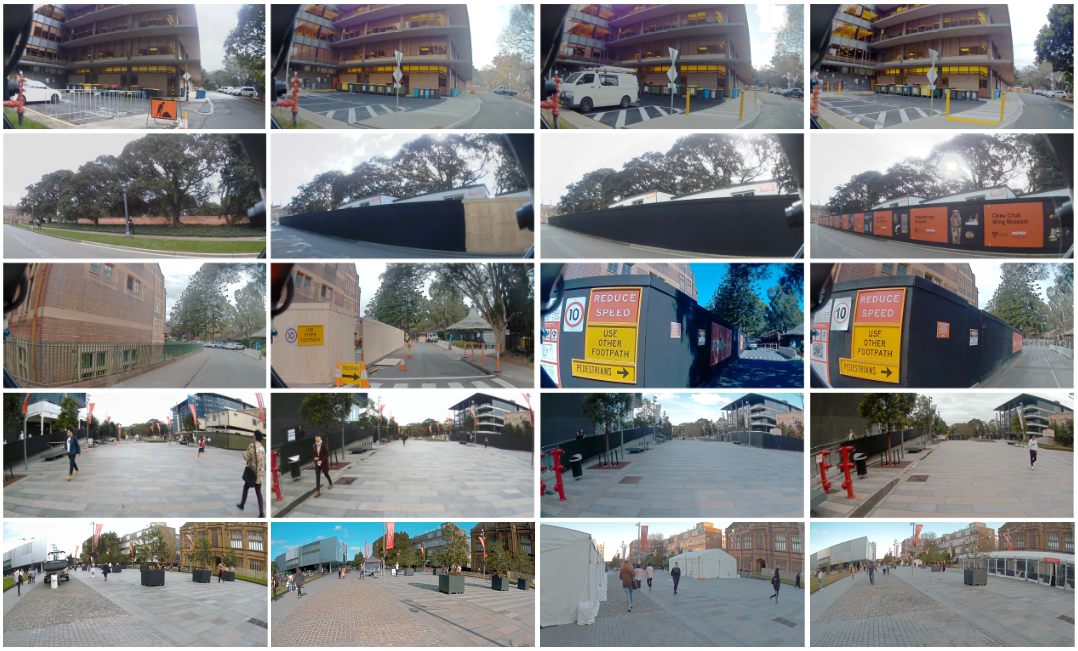}
    \caption{\small Structural changes discovered in the USyd Dataset. Each row of images shows a specific architectural/environmental transformation for each zone 1 through 5.}
    \label{fig:dataset_changes}
\end{figure*}

Fig. \ref{fig:dataset_changes} shows some of the structural changes that took place during the time of the dataset, which significant changes occurring in each of the different zones.

\subsection{Initial Feature Map}

The initial 2D feature-based map used as a prior for the map maintenance process was created using the data from the first dataset collected at the University of Sydney campus. The prior map consists of features obtained from a labelled point cloud \cite{stephany_iros} with its position optimised using a graph-based SLAM algorithm.

\subsubsection{Labelled point cloud}

We added semantic labels to the lidar point cloud by following the approach presented in \cite{stephany_journal_2020}. The cameras on the vehicle were intrinsically calibrated using a modified version of the \textit{camera calibration} ROS package \cite{ros_camera_calib}, which was changed to use a generic fish eye camera model \cite{fish_eye_model}. The extrinsic parameters of the lidar-camera pairs were calibrated as specified in \cite{SurabhiITSC}.

Images of the environment are processed by a CNN which outputs a semantic label for each pixel. The CNN discriminates between twelve different semantic classes: \textit{pole, building, road, vegetation, undrivable road, pedestrian, rider,  sky, fence, and vehicle} and \textit{unknown} (unlabelled and void).
As the lidar is mechanically scanned, we apply a correction to the lidar point cloud to compensate for the vehicle's motion. 
This correction also is constructed to align the lidar scan to the timestamp of the camera images. 
We then transfer the semantic information from the labelled images to the lidar point cloud using the camera calibration parameters, and the geometric relationship with the lidar. During this process, a masking technique is performed to avoid labelling lidar points which are not visible to the camera \cite{stephany_journal_2020}. 

\subsubsection{Feature extraction}

The features used for the mapping process are extracted from a semantic labelled lidar point cloud. These features are classified as poles or corners based on their geometrical shape. Poles are defined as cylindrical structures with an established minimum height and maximum width. Corners, on the other hand, are characterized as a vertical stack of the intersections between detected straight line segments with similar properties. In the urban environment, the straight lines are often observed from lidar returns on the walls of buildings \cite{siqi_iv}. It is worth mentioning that for the initial construction of the map, the parameters for the feature detectors were set to be very rigid in order to ensure that the features included in the initial map were stationary and increase the chance that the loop closure is successful. In our case, we consider valid poles to be taller than 1.8 meters with a diameter smaller than 0.3 meters, and valid corners are considered to be those taller than 1.6 meters.

\begin{figure}[h]
\vspace{3mm}
\centering

\begin{subfigure}[]{0.99\columnwidth}
\centering
	\includegraphics[width=\columnwidth]{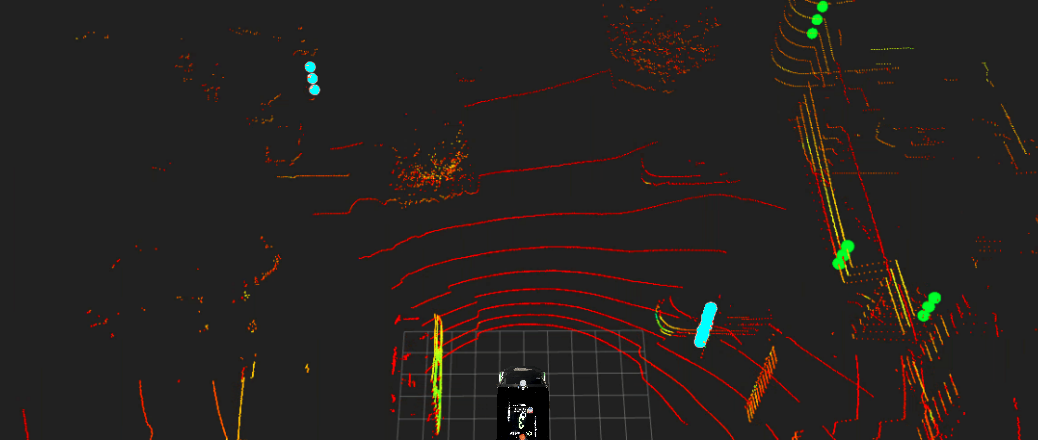}
    \caption{Pole and corner features detected from the point cloud.}
    \label{lidar_pc}
    \end{subfigure}
    
\begin{subfigure}[]{0.99\columnwidth}
\centering
	\includegraphics[width=\columnwidth]{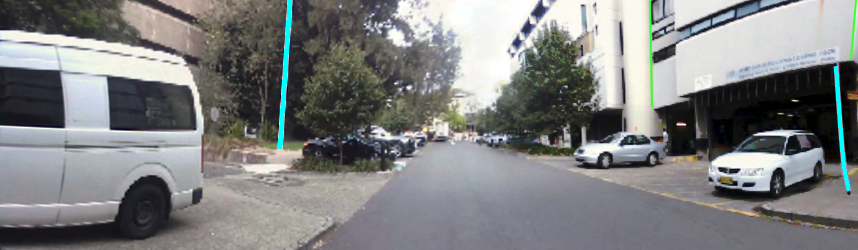}
    \caption{Front 160$^\circ$ FOV image from the surroundings. }
    \label{image_pc}
    \end{subfigure}
    
\caption{\small Features extracted from the lidar point cloud, poles in \colorbox{Cyan}{cyan}, corners in \colorbox{LimeGreen}{green}. }
\label{fig:pointcloud_features}
\end{figure}

Fig. \ref{fig:pointcloud_features} shows a example image/point cloud from the dataset with the features superimposed over the image. The features are constructed from stacks of vertical 3D points that are projected into the vehicle ground plane as a single 2D point (as if from a top down view). Every projected 2D point consists of: $(x,y)$ coordinates in the vehicle's frame, height, predominant semantic label, type of feature and a geometric property (diameter for poles and angle for corners).

The map building process uses a graph-based SLAM algorithm, integrating the detected lidar features, IMU, wheel encoders and GNSS data. An ICP data association method is used to find the correspondences between the different observations of lidar-based features. The GNSS information is used in two steps. The first step is to find the transformation matrix between the relative vehicle's pose and the global reference frame, to then apply the corresponding rotation and translation to the features' local position. In the second step, the vehicle's pose vertices are loosely constrained by the GNSS (due to its high uncertainty). Orthophotos are used to label poles and obtain their map-feature correspondence by a nearest neighbour algorithm. These relationships are incorporated as graph edges with tight constraints, and are added to the optimisation algorithm in order to adjust the features to the aerial image \cite{SiqiITSC}.

\subsection{Localisation algorithm}

As the primary purpose of the feature-based map is to localise the vehicle within it, we tested the map maintenance process through the localisation algorithm's performance.
The localisation algorithm is initialised using GNSS information. We used two successive GNSS readings to obtain a reasonable approximation of the vehicle's heading angle. These GNSS coordinates are only collected when the vehicle speed exceeds a certain threshold (more than \textcolor{red}{3} m/s, determined empirically but related to the noise of the GNSS). The initial vehicle heading angle is then set as the angle of the vector formed by these two geo-referenced points.
An unscented Kalman filter (UKF), integrates the GNSS, IMU and wheel encoders to estimate the position of the vehicle within a global frame. 

The existing features from within a 40 meter range of the estimated vehicle pose are retrieved from the prior map, and then compared to the currently observed features. 
Once the first successful map correspondence is found, the UKF updates the vehicle position using the additional observations from the ICP map matching algorithm.

\begin{figure}[b]
\centering
    \includegraphics[width=0.9\columnwidth]{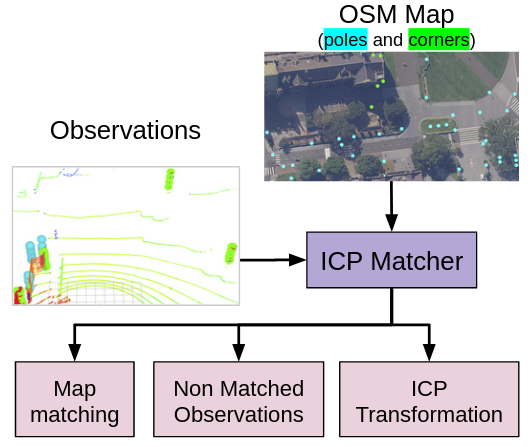}
    \caption{\small ICP matcher's outputs. }
    \label{fig:icp_matcher}
\end{figure}

An ICP matching algorithm is used for the data association task. It provides the UKF with a transformation between the current observations and the globally referenced features from the prior map that were close to the vehicle's estimated position.
In addition to the ICP transformation, the output of the matching process includes whether each feature was matched or not. This information is needed for the long-term map maintenance pipeline. 
The outputs from the ICP matcher consists of two vectors of features as shown in Fig. \ref{fig:icp_matcher}. The first vector includes the features that have been successfully matched against the observations.  The second vector contains the observed features which were not associated to any features within the map. 
Non-matched observations are the pole and corner features detected by the vehicle sensors which are not included in the localisation prior map.
This information is particularly relevant when considering adding new features into the map.  
For both the matched and non-matched feature observations, the output vectors include position, feature type, semantic label, height and geometric parameters. This data is useful to update existing, and initialise new features.

\begin{figure}[h]
\centering
    \includegraphics[width=0.9\columnwidth]{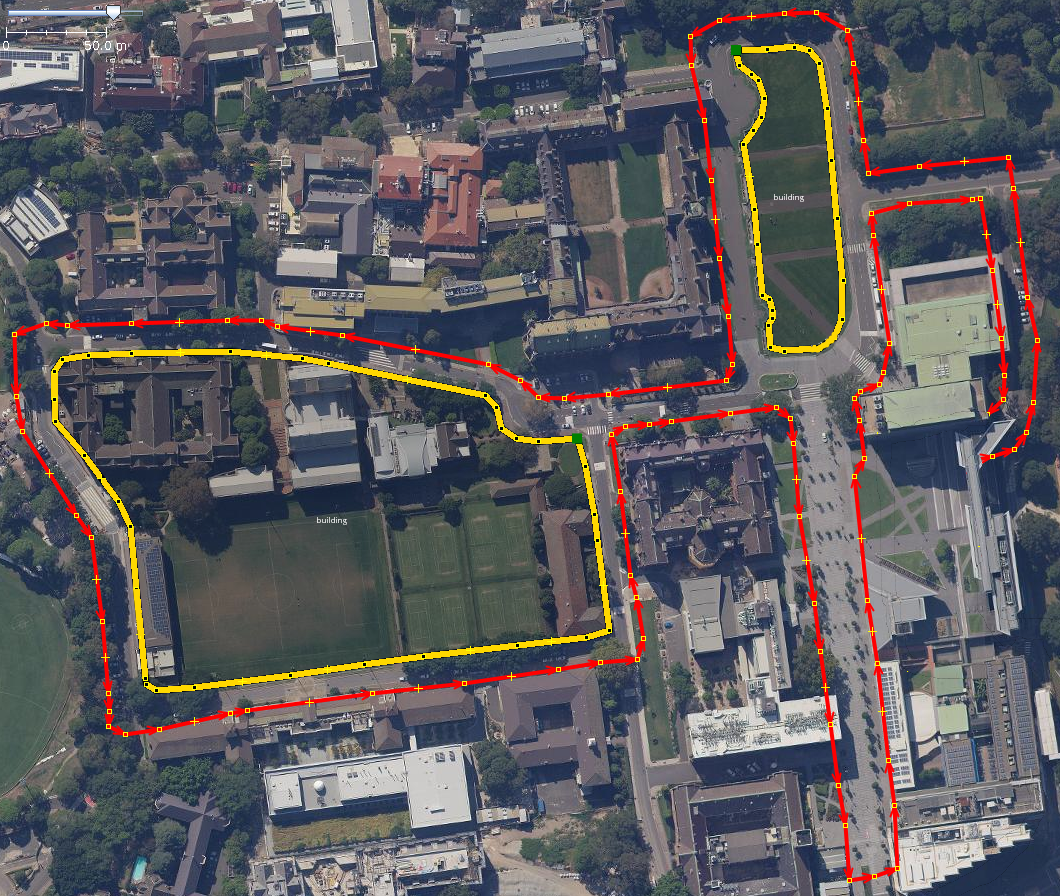}
    \caption{\small Valid drivable areas. The area enclosed with red boundaries (excluding the building area in yellow) corresponds to the valid roads visited during the collection of the dataset. }
    \label{fig:areas}
\end{figure}

The lack of features in certain areas, or an incorrect data association can have catastrophic consequences in an operational platform. 
Given that we know beforehand the areas where the vehicle will be operating, we can mitigate this problem by setting boundaries to indicate the valid drivable space. The drivable zones are specified as lanelet areas which enclose the roads visited by the vehicle in the dataset that do not belong to buildings or other undrivable areas (as parks or soccer fields). Fig. \ref{fig:areas} shows an illustration of the lanelets used to represent the roads visited in our dataset, the red and black areas are used to indicate an invalid position and cause the UKF module to be reset if the vehicle pose estimate is within these areas.
A reset signal is used to trigger the UKF module to re-initialise all variables and reset the vehicle's uncertainty estimate.

\section{Experiments and Results}

This section presents a number of experimental results that demonstrate the functionality of our proposed pipeline using the evaluation modules introduced in the previous section. We ran the proposed pipeline, and the localisation module over six months of data from the USyd Campus Dataset \cite{Usyd_dataset}. 
For the first week's data, we constructed the prior map as described in the previous section. This resulted in a feature map including 405 pole and corner features. 
From the second week, the map maintenance pipeline was run. 
During the experiment, we have also reduced the restrictions of the feature detector algorithm. The parameter affected in this case is the minimum height of each feature. We did this to be able to detect new features which can be beneficial for the localisation algorithm.
After that, we run the algorithm every 3 weeks, in each of these weeks we evaluated the number of new features included in the localisation map and the number of features removed from it.

\begin{table}[t]
\centering
\caption{\small Number of features}
\label{table:n_ft}
\begin{tabular}{|c|c|c|c|c|c|c|}
\hline
\multirow{2}{*}{W} & \multicolumn{2}{c|}{Height parameter} & \multirow{2}{*}{Reset} & \multirow{2}{*}{N. Features} & \multirow{2}{*}{R. Features} & \multirow{2}{*}{Total} \\ \cline{2-3}
                      & Pole            & Corner            &                        &                              &                              &                        \\ \hline
1                     & 1.6              & 1.8                & -                      & 405                          & -                            & 405                    \\ \hline
2                     & 1.6              & 1.8                & 3                      & 20                             &   22                           & 403                       \\ \hline
5                     & 1.6              & 1.8                & 2                      & 10                             &  5                            &   408                     \\ \hline
8                     & 1.6              & 1.5                & 2                      & 8                             &   2                           &  414                      \\ \hline
11                    & 1.5              & 1.5                & 0                      & 17                             &   6                           & 425                       \\ \hline
15                    & 1.4              & 1.5                & 0                      & 9                             &   5                           & 429                       \\ \hline
18                    & 1.4              & 1.4                & 0                      & 17                             &  3                            & 443                       \\ \hline
21                    & 1.4              & 1.4                & 0                      &  6                            &  2                            & 449                       \\ \hline
24                    & 1.4              & 1.4                & 0                      & 5                             &  2                            & 452                       \\ \hline
\end{tabular}
\label{tab:results}
\end{table}

Table \ref{tab:results} shows the number of new and removed features (denoted as N. and R. Features respectively in the table). The total number of elements in the map is displayed in the last column. For every week in the first column, the minimum height parameter of the feature detector is listed.
In the fourth column, we indicate the number of times the localisation filter was reset during that specific week. 
We use the number of times the filter resets as a proxy to measure the ability of the localisation filter to work given a specific prior feature map.
During weeks 2, 5 and 8, two or more reset signals were generated, this occurred in areas with small numbers of detectable features. These regions did not have strong feature patterns in these locations, which resulted in either incorrect data association or a large accumulation of error from the odometry. 
After week 11, there were sufficient features included in the map due to the relaxation of the height parameter for the feature detector. 
The additional features in previously low density areas resulting from this change improved the localisation resulted in no further resetting of the filter. 


\begin{figure}[b]
\centering
    \includegraphics[width=\columnwidth]{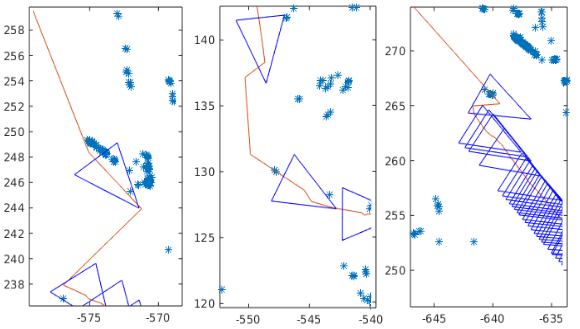}
    \caption{\small Strong corrections in positioning.}
    \label{fig:discontinue}
\end{figure}

Another metric we use to measure the quality of the map for localisation is the number of large position corrections due to the update step of the filter. 
There are a number of reasons why there would be large corrections related to map and sensor noise.
When the vehicle is unable to feature match for a period of time, the error caused by dead reckoning increases over time. The discontinuity in position estimate is more likely to occur when a successful match takes place after a period of time with no matches.
Incorrect data association, or incorrectly positioned features within the map can also cause the previously discussed behaviour.
For the experiment, we count the number of strong corrections to demonstrate the ability to match observations to a specific map.
We define strong corrections as changes in the position with a large lateral translation, which is not physically possible due to holonomic constraints. 
Fig. \ref{fig:discontinue} shows some examples of discontinuities in the vehicle trajectory.

\begin{figure}[t]
\centering
    \includegraphics[width=\columnwidth]{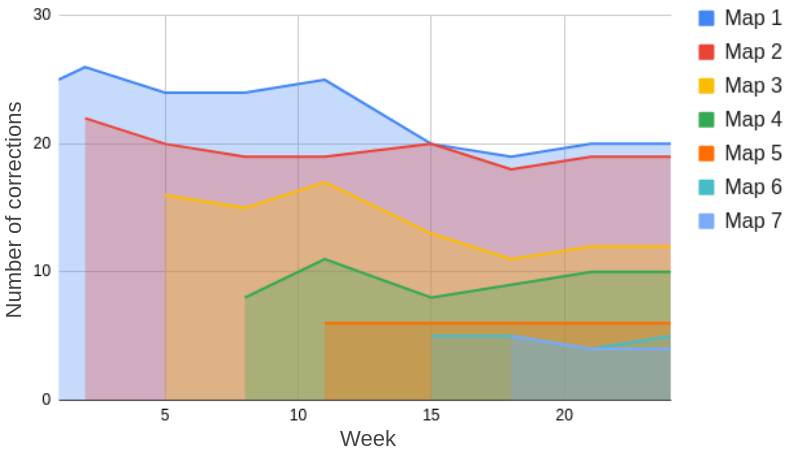}
    \caption{\small Number of strong corrections per updated map. The number of weeks lies on the x-axis while the number of corrections is set in y-axis.}
    \label{fig:map_graph}
\end{figure}

After updating the map, we run the localisation algorithm to localise the vehicle within the updated map in the subsequent weeks. 
Fig. \ref{fig:map_graph} shows the number of times a strong correction occurred per experiment per map. 
In this figure, map 2 is the updated version of map 1; map 3 is the updated version of map 2 and so on. 
We can see that the number of corrections across datasets using the same map are relatively stable, and for some datasets, it is even lower than the initial value. The reduction of the number of strong corrections are mainly due to the change of height parameter for the feature extractor. 
When the vehicle is able to identify shorter poles and corners, the features within the map can also be detected from a larger distance, generating more updates and smaller corrections.
Overall, the performance of each updated map shows an improvement on the previous map across the dataset. 
This can be seen in Fig. \ref{fig:map_graph} that all graphs generated by updated maps have a consistent smaller number of strong corrections.

\begin{figure}[h]
\centering
     \begin{subfigure}[b]{0.48\columnwidth}
    \centering
    \includegraphics[width=\textwidth]{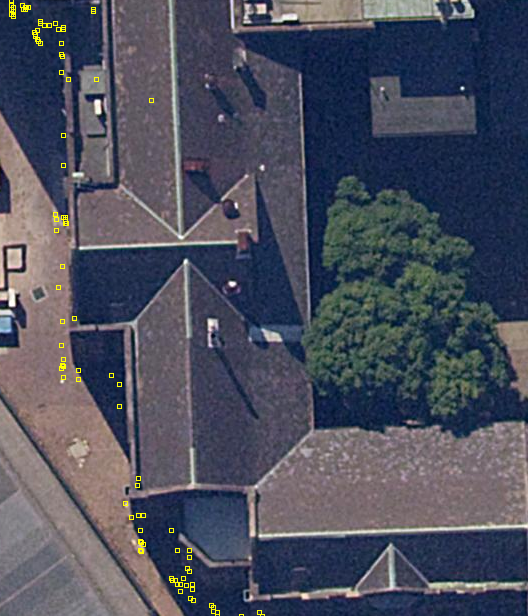}
    \caption{Week 2}
    \label{fig:week2}
  \end{subfigure}
    \begin{subfigure}[b]{0.48\columnwidth}
    \centering
    \includegraphics[width=\textwidth]{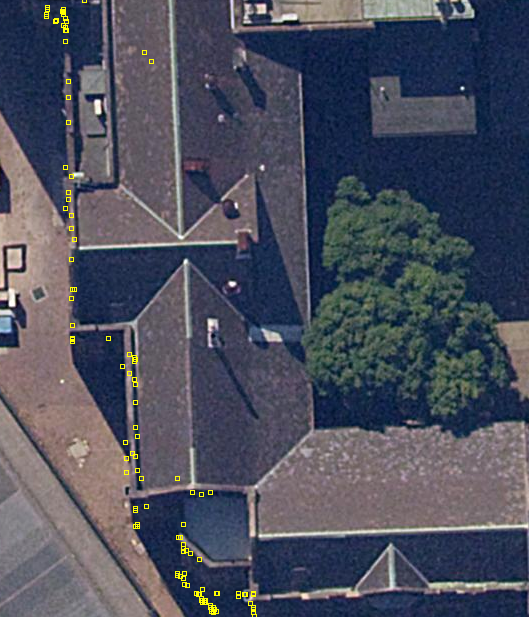}
    \caption{Week 5}
    \label{fig:week5}
  \end{subfigure}
  \begin{subfigure}[b]{0.48\columnwidth}
  \centering
    \includegraphics[width=\textwidth]{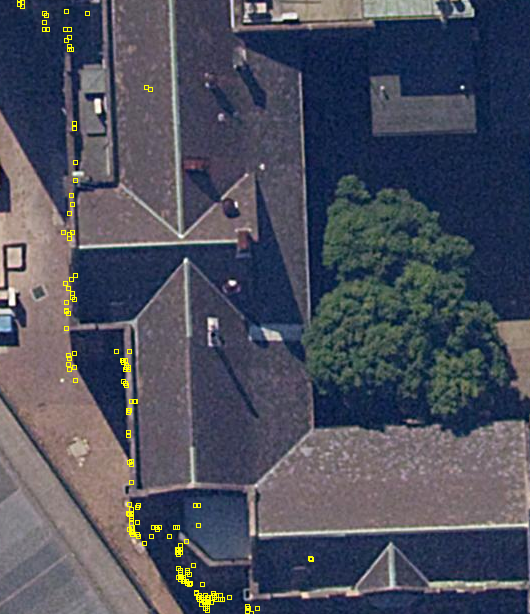}
    \caption{Week 8}
    \label{fig:week8}
  \end{subfigure}
  \centering
  \begin{subfigure}[b]{0.48\columnwidth}
    \includegraphics[width=\textwidth]{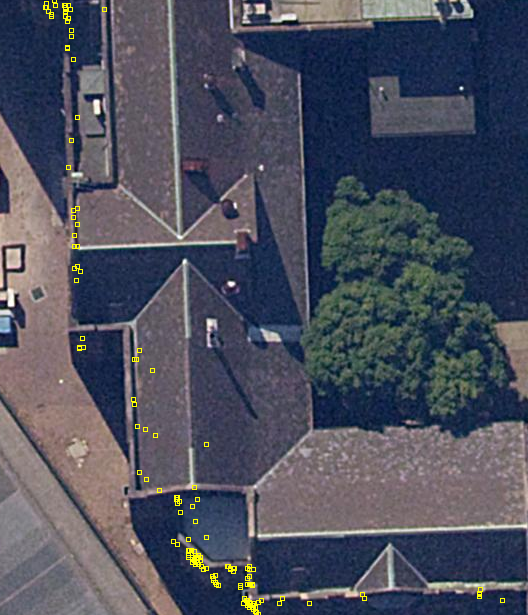}
    \caption{Week 11}
    \label{fig:week11}
  \end{subfigure}
  \begin{subfigure}[b]{0.48\columnwidth}
  \centering
    \includegraphics[width=\textwidth]{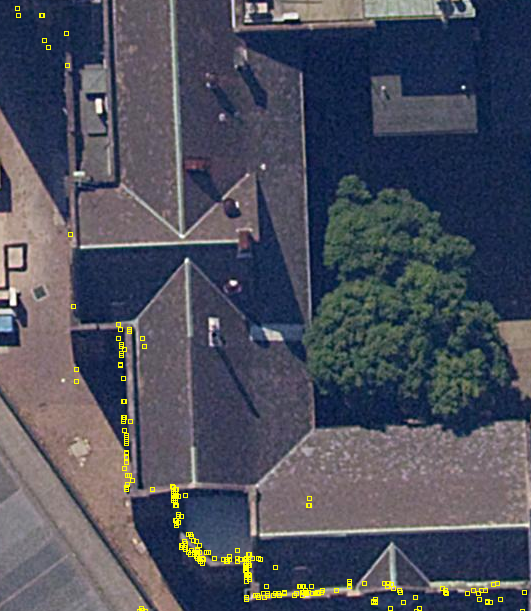}
    \caption{Week 15}
    \label{fig:week15}
  \end{subfigure}
  \centering
  \begin{subfigure}[b]{0.48\columnwidth}
    \includegraphics[width=\textwidth]{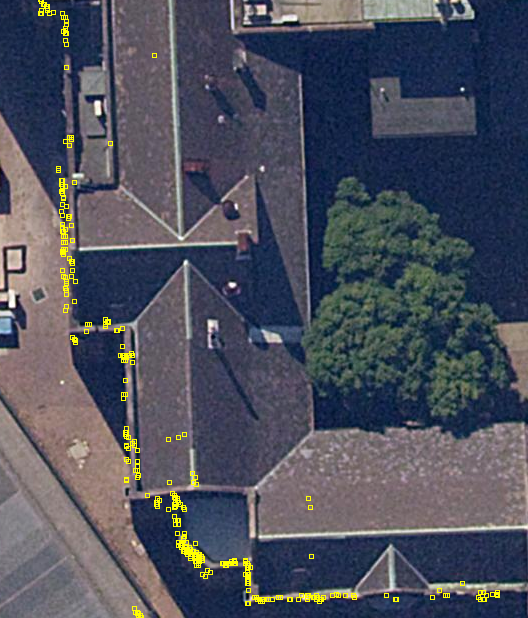}
    \caption{Week 18}
    \label{fig:week18}
  \end{subfigure}
    \caption{\small Evolution of the overlap between detected features and satellite orthophotos within an initially featureless area. }
    \label{fig:map_overlap}
\end{figure}

Qualitative results are shown in Fig. \ref{fig:map_overlap}. In this figure, we overlay all of the detected features on an orthophoto of the area. 
Initially, the area contains few features to match against, causing lower quality feature matches and increased reliance on dead reckoning. 
In Fig. \ref{fig:week2}, \ref{fig:week5} and \ref{fig:week8} we can see that a number of detected features (in yellow) are shifted with a bias towards the bottom of the image. 
This deviation is in the vehicles direction of travel and is a consequence of the odometry errors accumulated over time. After Fig. \ref{fig:week11} a number of new features were detected and included in the map. The localisation algorithm was able to match more observations against the updated map. As a result of this, the localisation filter benefits from the improved output of the data association algorithm to estimate a more accurate position. By reducing the localisation error, the observed features translated into a global reference frame have an improved overlap with the orthoimagery of the area.  
As more features are included in the map, the localisation become more accurate.  Fig. \ref{fig:week15} and \ref{fig:week18} demonstrate the improvement of the localisation in terms of the correspondence between the detected features and the reference orthophoto.


\begin{figure}[t]
\centering
    \includegraphics[width=\columnwidth]{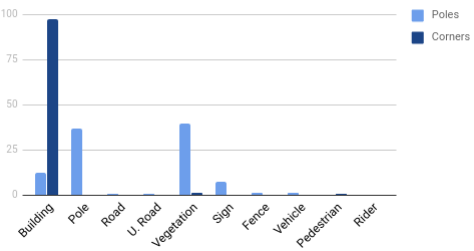}
    \caption{\small Labels in poles and corners.}
    \label{fig:histo}
\end{figure}

Finally, we use the semantically labelled point cloud to generate the transformations between labelled images and the lidar point cloud. 
Fig. \ref{fig:histo} shows the normalised histogram of the semantic labels for each type of feature. 
As expected, the corner features were mostly detected correctly from structures identified as buildings in the labelled image. 
Pole features were mostly labelled as either pole and vegetation with a few outlier labels. This is because many of the pole features were from the trunks of trees.

\section{Conclusions and Future work}
In this paper, we have presented a pipeline for long-term maintenance of feature-based maps. Our approach is based on a layered structure, where a prior map layer contains the positions and attributes of each feature. These attributes include a measure of the visibility of each map feature. This visibility vector allows us to characterise the locations from which the vehicle has previously detected the features, and the frequency of these observations. 
Subsequent visits to the same locations are used to re-evaluate the visibility vector; determining how often and from where the features are re-observed.
The visibility volume metric presented in this paper is used to assess the stability of the features. 
By evaluating this metric we are able to remove features that no longer exist from the next iteration of the prior map layer. 

We built and maintained a new feature map layer from the observations that could not be matched against the prior map layer.  We selected the most relevant features and adjusted them based on a graph optimisation algorithm. Moreover, we evaluate the density of the features in the different areas through the concentration ratio metric. This procedure allowed us to incorporate the newly detected features into the next iteration of the prior map layer while improving the distribution of map elements.

The pipeline implementation was tested using a long-term dataset containing significant changes to the operating environment over time. The performance was validated using a localisation algorithm receiving updates from map matching against the prior feature-based map. 

A comprehensive set of experiments were presented using a dataset incorporating several weeks of data. This data includes significant variation in the environment resulting from changes to buildings and other structures over the timeframe of the dataset.  A number of metrics were presented to evaluate quantitative results showing the improvement of localisation due to the map maintenance process. The results show the number of new features incorporated into the map, and the number of older features that were deleted based on the presented pipeline.
 
Qualitative results were shown by registering the detected features based on the vehicle's position and overlaying them on an orthoimage of the environment. The correspondence between these lidar features and orthoimage improves with the updates to the map. 

We used a semantically labelled point cloud as an input to the feature extractor in order to analyse the distribution of the feature labels in the map. As future work, we intend to incorporate the semantic labels into the pipeline as a new variable to discriminate between static and dynamic objects.


%

\section*{Acknowledgment}
This work has been funded by the ACFR, the University of Sydney through the Dean of Engineering and Information Technologies PhD Scholarship (South America) and University of Michigan / Ford Motors Company Contract ``Next generation Vehicles".

\ifCLASSOPTIONcaptionsoff
  \newpage
\fi

\bibliography{main}
\bibliographystyle{IEEEtran}

\begin{IEEEbiography}[{\includegraphics[width=1in,height=1.25in,clip,keepaspectratio]{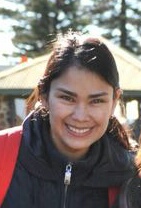}}]{Julie Stephany Berrio} received the B.S. degree in Mechatronics Engineering in 2009 from Universidad Autonoma de Occidente, Cali, Colombia, and the M.E. degree in 2012 from the Universidad del Valle, Cali, Colombia. She is currently working towards the Ph.D. degree at the University of Sydney, Sydney, Australia. Her research interest includes semantic mapping, long-term map maintenance, and point cloud processing.
\end{IEEEbiography}

\begin{IEEEbiography}[{\includegraphics[width=1in,height=1.25in,clip,keepaspectratio]{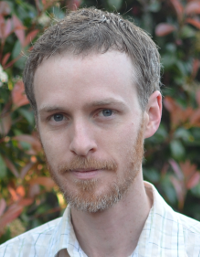}}]{Stewart Worrall} received the Ph.D. from the University of Sydney, Australia, in 2009. He is currently a Research Fellow with the Australian Centre for Field Robotics, University of Sydney. His research is focused on the study and application of technology for vehicle automation and improving safety.
\end{IEEEbiography}

\begin{IEEEbiography}[{\includegraphics[width=1in,height=1.25in,clip,keepaspectratio]{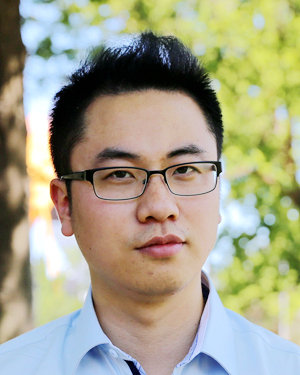}}]{Mao Shan} received the B.S. degree in electrical engineering from the Shaanxi University of Science and Technology, Xi’an, China, in 2006, and the M.S. degree in automation and manufacturing systems and Ph.D. degree from the University of Sydney, Australia, in 2009 and 2014, respectively. He is currently a Research Fellow with the Australian Centre for Field Robotics, the University of Sydney. His research interests include autonomous systems, localisation, and tracking algorithms and applications.
\end{IEEEbiography}

\begin{IEEEbiography}[{\includegraphics[width=1in,height=1.25in,clip,keepaspectratio]{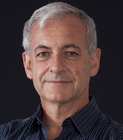}}]{Eduardo Nebot} received the BSc. degree in electrical engineering from the Universidad Nacional del Sur, Argentina, M.Sc. and Ph.D. degrees from Colorado State University, Colorado, USA. He is currently a Professor at the University of Sydney, Sydney, Australia, and the Director of the Australian Centre for Field Robotics. His main research interests are in field robotics automation and intelligent transport systems. The major impact of his fundamental research is in autonomous systems, navigation, and safety.
\end{IEEEbiography}





\end{document}